\documentclass[lettersize,journal]{IEEEtran}
\usepackage{algorithmic}
\usepackage{algorithm}
\usepackage{array}
\usepackage[caption=false,font=normalsize,labelfont=sf,textfont=sf]{subfig}
\usepackage{textcomp}
\usepackage{stfloats}
\usepackage{url}
\usepackage{verbatim}
\usepackage{graphicx}
\usepackage{subcaption}
\usepackage{cite}
\usepackage[dvipsnames]{xcolor}
\usepackage{amsmath,amsfonts}
\usepackage{amssymb}

\begin{document}

\title{Learning Latent Representations for Image Translation using Frequency Distributed CycleGAN}

\author{Shivangi Nigam, Adarsh Prasad Behera, \textit{Member, IEEE},
        Shekhar Verma, \textit{Senior member, IEEE}, and P. Nagabhushan, \textit{Senior member, IEEE}
\thanks{Shivangi Nigam and Shekhar Verma are with the Department of Information Technology, Indian Institute of  Information Technology, Allahabad, U.P. , 211012 India, e-mail:\{rsi2018506, sverma\}@iiita.ac.in}
\thanks{Adarsh Prasad Behera is with KTH Royal Institute of Technology, 100 44 Stockholm, Sweden, e-mail: apbehera@kth.se}
\thanks{P. Nagabhushan is with the Department of Computer Science and Engineering, Vignan University, Guntur, Andhra Pradesh 522213, India e-mail: pnagabhushan@iiita.ac.in}}

\markboth{IEEE Transactions on Image Processing }%
{Shell \MakeLowercase{\textit{et al.}}: A Sample Article Using IEEEtran.cls for IEEE Journals}


\maketitle

\begin{abstract}
This paper presents Fd-CycleGAN, an image-to-image (I2I) translation framework that enhances latent representation learning to approximate real data distributions. Building upon the foundation of CycleGAN, our approach integrates Local Neighborhood Encoding (LNE) and Frequency-aware supervision to capture fine-grained local pixel semantics while preserving structural coherence from the source domain. We employ distribution-based loss metrics, including KL/JS Divergence and log-based similarity measures, to explicitly quantify the alignment between real and generated image distributions in both spatial and frequency domains. To validate the efficacy of Fd-CycleGAN, we conduct experiments on diverse datasets—Horse2Zebra, Monet2Photo, and a synthetically augmented Strike-off dataset. Compared to baseline CycleGAN and other state-of-the-art methods, our approach demonstrates superior perceptual quality, faster convergence, and improved mode diversity, particularly in low-data regimes. By effectively capturing local and global distribution characteristics, Fd-CycleGAN achieves more visually coherent and semantically consistent translations. Our results suggest that frequency-guided latent learning significantly improves generalization in image translation tasks, with promising applications in document restoration, artistic style transfer, and medical image synthesis. We also provide comparative insights with diffusion-based generative models, highlighting the advantages of our lightweight adversarial approach in terms of training efficiency and qualitative output.

\end{abstract}


\section{Introduction}

\begin{figure}[t]
    \centering
    \includegraphics[scale=0.55]{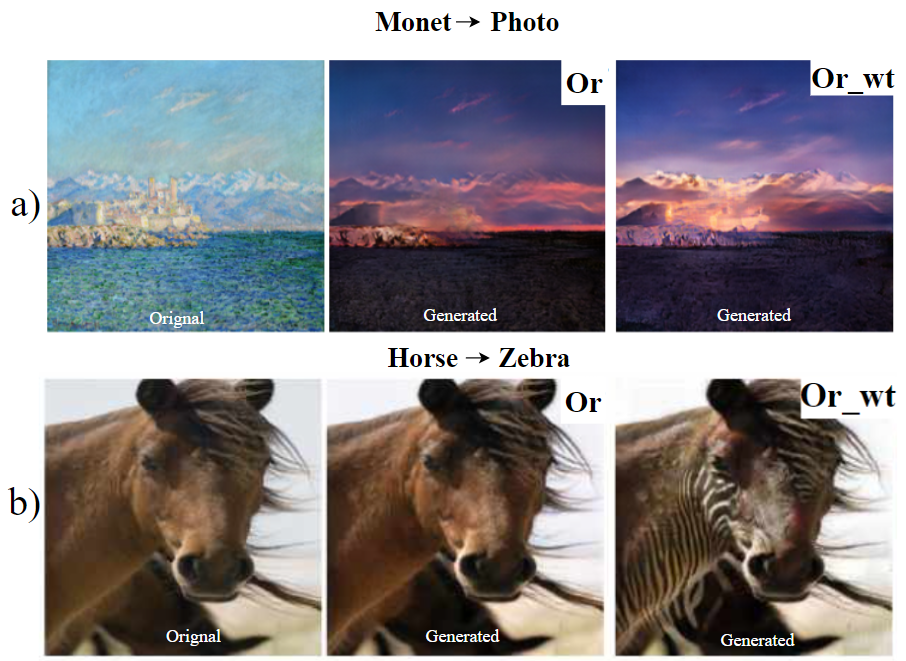}
    \caption{An example of the difference between conversion from CycleGAN and Fd-CycleGAN }
    \label{fig:sample}
\end{figure}

Domain Translation (DT), also referred to as I2I translation, involves learning a mapping between two visual domains, often in the absence of paired data. This task has become central to several vision tasks. For these tasks, generative models have been widely adopted \cite{cao2023DT, couairon2022flexit_DT, shamsolmoali2022gen_DT}. These models generate synthetic data through a stochastic process $z \sim p(z)$, typically drawn from a simple prior, is transformed into an observed data sample $x \sim p_{\theta} (x\mid z)$ using a generator $\mathcal{G}(z;\theta_g)$. The objective is to approximate the true data distribution by maximizing the data likelihood $\pi_{i=1}^{n} p_{\theta}(x_i)$. Explicit density models, such as MLE and Markov Chain-based methods \cite{arora2017gans_dist}, explicitly define a marginal likelihood $p_{\theta}x$. GANs \cite{saxena2021generative_GAN_review}, do not require an explicit likelihood function.However, GANs pose significant training challenges \cite{gui2021review_GAN}, primarily due to the non-convexity of the minimax objective, instability in convergence, and issues like mode collapse \cite{hong2019generative_GANreview, kossale2022mode_mode_coll}, reducing diversity in the output. Strategies such as Wasserstein GAN (WGAN) with Lipschitz continuity \cite{arjovsky2017wassersteinGAN} and spectral normalization stabilize training and improve latent space representation learning. In the context of Unpaired Domain Translation (UDT), the goal is to learn mappings between two unpaired domains $\mathcal{X},\mathcal{Y} \subseteq \mathbb{R}^d$  via transformations $\mathcal{T}: \mathcal{X} \to \mathcal{Y}$ and 
$\mathcal{F}: \mathcal{Y} \to \mathcal{X}$, underpinned by their respective distributions. Most UDT approaches enforce cycle-consistency and adversarial coherence to constrain the solution space. However, CycleGAN often suffer from semantic ambiguity, mode collapse, and loss of local structure, especially in complex scenes where accurate object-background separation is critical. CycleGAN attempts to learn the joint distribution between source and target domains using only marginal distributions from unpaired data. The adversarial loss facilitates the learning of this joint distribution by setting up a competitive dynamic between the generator and discriminator. However, this formulation results in an ill-posed problem since multiple joint distributions can correspond to the same marginals. Without explicit constraints, the solution space remains ambiguous.

Recent methods such as CUT \cite{CUT2020contrastive} and diffusion-based models \cite{Diffusion_1_2020denoising, diffusion_2rombach2022high} have significantly improved I2I translation by leveraging contrastive learning and probabilistic modeling. However, these methods often demand high computational resources or rely on substantial supervision. In parallel, increasing attention has been given to frequency and structural domain modeling to mitigate semantic loss and enhance content preservation. Wavelet-domain GANs \cite{wavelet_based_eskandar2023waveletbasedunsupervisedlabeltoimagetranslation} and frequency-structure-aware translation methods \cite{liu2023frequency-aware-modelling} incorporates spectral features to retain textural fidelity during transformation. Spectral Normalization with Dual Contrastive Regularization (SN-DCR) \cite{SN-DCN_zhao2024spectralnormalizationdualcontrastive} and Implicit Bridge Consistency Distillation (IBCD) \cite{IBCD_2025single}, adopt contrastive or distillation-based mechanisms to promote structural alignment without relying on adversarial learning.

Our work introduces a novel frequency-aware preprocessing module that can be seamlessly integrated into any CycleGAN variant without architectural modification, thereby preserving computational efficiency and generalizability. By combining global frequency divergence with localized spatial constraints through neighborhood encoding, our framework simultaneously captures semantic coherence and fine-grained details. Our method demonstrates robust performance across diverse datasets. Our enhancements to the CycleGAN framework—frequency distribution modeling and local neighborhood encoding—lead to richer latent representations that preserve pixel-level semantic coherence. The key contributions of this work are summarized as follows:

\begin{enumerate}
    \item We propose \textit{Fd-CycleGAN}, an enhanced translation framework that integrates LNE and frequency-aware functions (Fd) into CycleGAN for improved semantic and structural preservation.
    \begin{itemize}
        \item \textit{LNE} encodes local similarity by embedding neighborhood features into each input pixel to inform the discriminator and generator during adversarial training.
        \item \textit{Fd} models learn frequency priors of the data to capture both global and local semantic structures essential for accurate translation.
    \end{itemize}
    
    \item We design new loss functions—Divergence and Log-based metrics—that replace the L1 norm in vanilla CycleGAN to enforce cycle consistency through a distribution-sensitive lens.

    \item We conduct comprehensive evaluations across diverse datasets to examine the generalizability, performance bounds, and optimal configurations of Fd-CycleGAN under varied frequency and semantic structures.
\end{enumerate}

\section{Related work}\label{sec:lit_rev}

In this section, we survey foundational models and recent advancements in the field \textit{Fd-CycleGAN}.

\begin{figure*}[ht]
    \centering
    \includegraphics[scale=0.5]{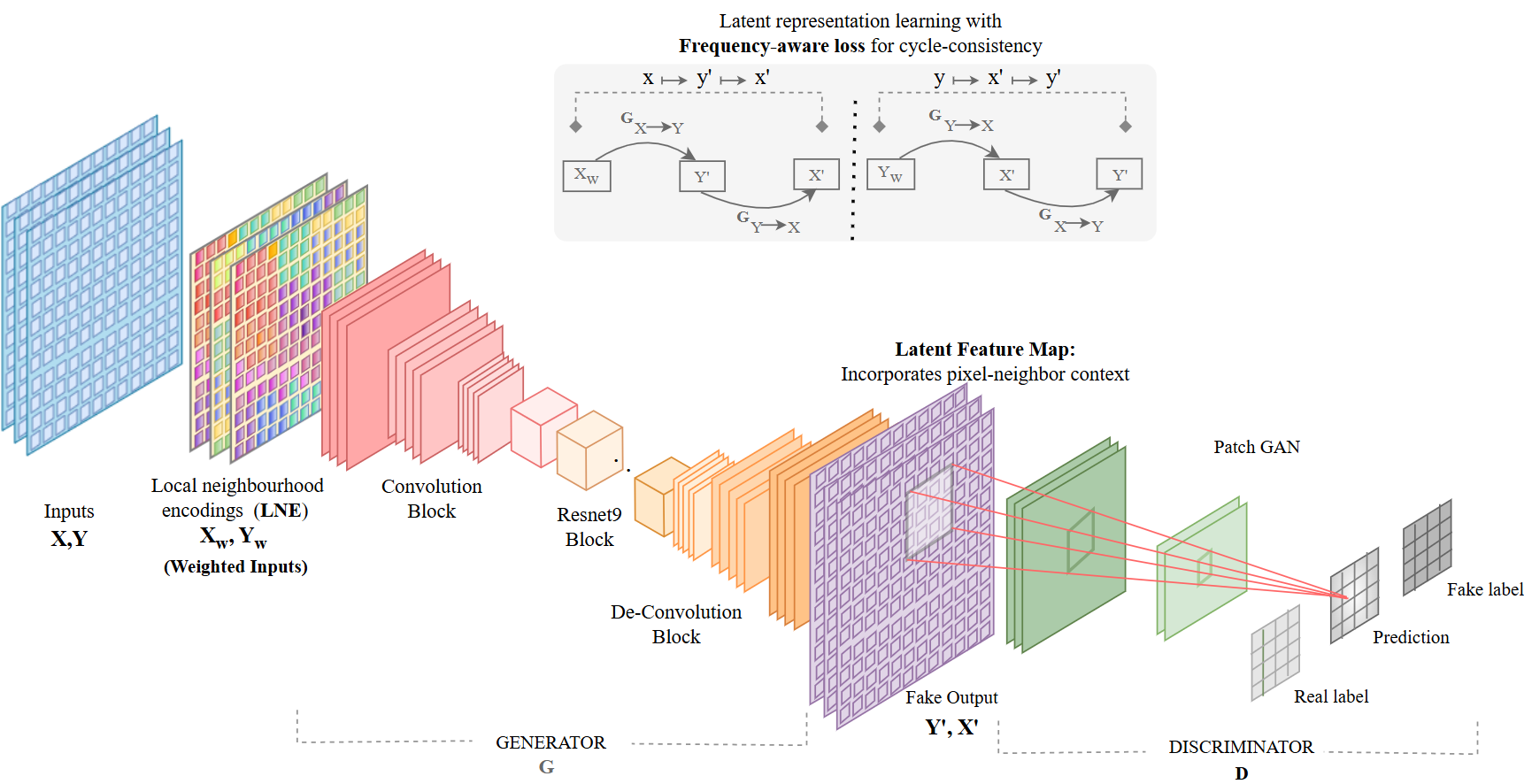}
    \caption{Overview of the proposed Fd-CycleGAN architecture. The input images are first enhanced using LNE, which encodes each pixel based on the spectral similarity with its two-hop neighboring pixels. This similarity is modeled using a Gaussian weighting function, resulting in an input representation enriched with local spatial context. The embedded inputs are then passed to the adversarial network. The Generator follows a ResNet-based architecture comprising nine residual blocks, along with initial convolution and final deconvolution layers. The Generators $\mathcal{G}_{x \rightarrow \ y}, \mathcal{G}_{y \rightarrow \ x}$ produce synthetic images which are evaluated by Dicriminators $\mathcal{D}_X, \mathcal{D}_Y$. The frequency-aware cycle-consistency loss captures the semantics of the data. }
    \label{fig:architecture}
\end{figure*}

\textbf{Paired and Unpaired I2I Translation} models like Pix2Pix \cite{isola2017Pix2Pix} pioneered supervised I2I translation using conditional GANs, requiring paired datasets, which limits its applicability. CycleGAN \cite{zhu2017_CycleGAN}, a seminal unpaired method, introduced cycle consistency to learn mappings between domains without one-to-one correspondence. CycleGAN often suffers from semantic distortions and texture loss, especially in domains where structural and frequency preservation is critical. CUT \cite{CUT2020contrastive} improves translation quality using patch-level contrastive learning but increases training complexity and requires additional memory overhead. Other extensions, such as MUNIT and DRIT, enable multimodal outputs but still lack explicit control over spatial or frequency features.

\textbf{Diffusion Models} offering superior perceptual quality through iterative denoising. Denoising Diffusion Probabilistic Models (DDPM) \cite{Diffusion_1_2020denoising}, Latent Diffusion \cite{diffusion_2rombach2022high}, and Palette \cite{saharia2022palette} have demonstrated remarkable results in high-resolution image synthesis. Applications of diffusion to I2I, such as Pix2Pix-Zero \cite{mokady2023pix2pixzero} and Plug-and-Play Diffusion \cite{choi2023plugplay}, bring flexibility but are computationally expensive.

\textbf{Frequency-Aware and Structural Preservation Techniques} have explored frequency-domain cues to preserve texture and structure.  \cite{wavelet_based_eskandar2023waveletbasedunsupervisedlabeltoimagetranslation} use wavelet decomposition to retain high-frequency details in unsupervised label-to-image translation. \cite{liu2023frequency-aware-modelling} introduce frequency separation attention in GANs to enhance region-specific focus. However, these models often require architectural redesigns or introduce computational complexity. Spectral Normalization with Dual Contrastive Regularization (SN-DCR) \cite{SN-DCN_zhao2024spectralnormalizationdualcontrastive} and Information Bottleneck Contrastive Distillation (IBCD) \cite{IBCD_2025single} avoid adversarial losses altogether by employing perceptual and contrastive constraints to preserve structure. While effective in specific domains like semantic segmentation, these approaches are task-specific and may not generalize to low-level handwritten degradation scenarios.

\textbf{Lightweight, Plug-and-Play, Modular enhancements}, such as Adaptive Instance Normalization (AdaIN) \cite{adain2017}, Spatially-Adaptive Normalization (SPADE) \cite{spade2019semantic}, and attention-injected GANs, have demonstrated effective plug-and-play conditioning mechanisms.

\textbf{Few-Shot and Domain-Specific Translation} methods like FUNIT \cite{liu2019fewshot} and SEGA leverage small amounts of data and semantic alignment to perform cross-domain translation. These are particularly relevant for domains like ours, where paired data for handwritten strike-off scenarios is scarce. Moreover, while most I2I literature focuses on stylized image domains (e.g., faces, landscapes), fewer works explore translation in textual or document modalities. Some recent works \cite{handwritingGAN2020, textstylebrush2021} demonstrate GAN-based handwriting generation, yet they do not address real-world noise such as strike-offs or ink bleed.

To address the limitations of existing I2I frameworks in retaining semantic fidelity and structural consistency in degraded handwritten text scenarios, we propose \textit{Fd-CycleGAN}, a lightweight, architecture-agnostic enhancement to CycleGAN. By combining a frequency divergence objective with localized spatial context encoding, the model enforces both global and neighborhood-level feature preservation. Unlike resource-heavy diffusion models or domain-specific semantic regularization techniques, Fd-CycleGAN generalizes well across diverse scripts and is particularly robust to noise patterns like strike-offs—demonstrating competitive performance without sacrificing training efficiency.

\section{Methodology} \label{sec:proposed_work}

We propose \textit{Fd-CycleGAN}, an improved version of CycleGAN designed to enhance the learning of local representations while retaining the essential characteristics of the source domain. Unlike CUT \cite{CUT2020contrastive}, which removes the need for backward translation through contrastive learning, our model preserves the bidirectional mapping constraint of CycleGAN. We enhance the translation process through frequency-spatial regularization, which enables better control over image semantics. Our method incorporates a frequency-based regularization technique that suppresses high-frequency noise during translation. This is particularly beneficial in applications like document restoration, where structural consistency and cleanliness are essential.
Our approach facilitates smooth and high-fidelity I2I translation while preserving essential structural details from the source domain, such as textures and lines. Specifically, Fd-CycleGAN introduces two key enhancements over the original CycleGAN:

\begin{enumerate}
    \item \textbf{Local Neighborhood Encoding (LNE)}: This module encodes neighborhood similarity at the pixel level, allowing the generator to capture fine-grained spatial features effectively. LNE guides the generation process by enhancing spatial encoding and feature extraction.
    \item \textbf{Frequency-aware Similarity Computation}: These functions model latent representations based on both local and global semantic information, supporting robust generation across diverse image domains. These functions capture the cost of translation by learning meaningful latent distributions.
\end{enumerate}

Additionally, we introduce two loss metrics—Divergence-based and Log-based similarity measures—to quantify the alignment between generated and real images. These are incorporated within the cycle-consistency loss to improve convergence and maintain semantic consistency.

\subsection{Learning Latent representations of local neighborhood}

\subsubsection{\textbf{LNE}}
is a pre-processing step designed to encode the neighborhood similarity into each pixel of the input images which are then processed by the adversarial network. This technique has been effectively applied to various image processing tasks such as image super-resolution, denoising, and inpainting. The primary objective of LNE is to reduce noise and smoothen images while preserving the semantic structure of the local neighborhood. Before feeding an image into the network, it is transformed into a neighborhood-encoded representation by computing the Gaussian weights of neighboring pixels. The weighted images serves for two objectives: 

\begin{figure}[t]
    \centering
    \includegraphics[scale=0.48]{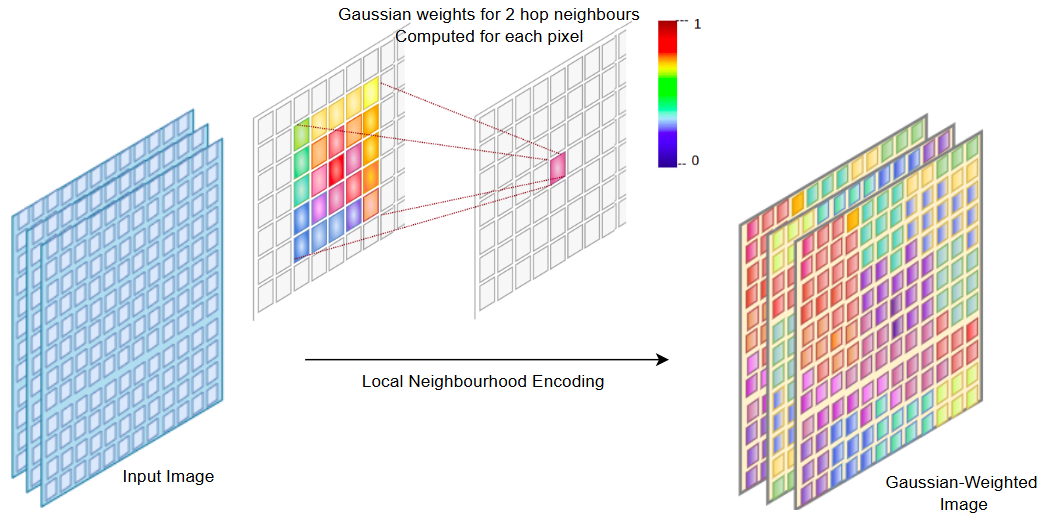}
    \caption{Illustration of the LNE mechanism. For each pixel in the input image, LNE computes a weighted representation by considering the spectral similarity with its surrounding neighbors within a two-hop radius. This results in a transformed image where each pixel encodes local structural context.}
    \label{fig:LNE}
\end{figure}

\begin{itemize}
    \item \textit{Preserving local semantics:} The Gaussian-based weighted image is based on the exponential decay of intensity differences between neighboring pixels. A lower weight is assigned to distant or dissimilar neighbors, while a higher weight is given to closer and more similar ones. This approach helps capture the spatial structure of the image, enabling the generator to produce more semantically consistent outputs.
    
    \item \textit{Enhancing image quality and reducing noise:} By assigning smooth, continuous weights to neighboring pixels, abrupt intensity variations are reduced. This leads to improved texture continuity, smoother image surfaces, and cleaner image reconstructions with more prominent and stable features.
\end{itemize}

The high-dimensional data sequence $\mathcal{X} = \{x_1, x_2, ..., x_n\}$, where $\mathcal{D} \in \mathbb{R}^H$, consists of input vectors with $H$ dimensions (columns) and $n$ instances (rows). The dissimilarity between two points $x_j$ and $x_i$ in $\mathcal{X}$ is modeled using conditional probabilities $p_{j|i}$. These probabilities are computed using a Gaussian distribution as follows:
\begin{equation}
    w_{j|i} = \exp\left(-\frac{\left\| x_i - x_j \right\|^2}{2\sigma_i^2} \right)
\end{equation}

Here, $w_{j|i}$ denotes the weight between pixels $i$ and $j$, with $\left\| x_i - x_j \right\|$ representing the Euclidean distance between them, indicating their similarity. The Gaussian distribution is centered at $x_i$, with variance $\sigma_i$, which controls the spread of the distribution. This formulation ensures smaller probabilities for distant neighbors and higher probabilities for nearby ones, effectively encoding spatial proximity. For each neighborhood, the weights are normalized by the sum of all weights within that neighborhood, yielding the conditional probability:
\begin{equation}
    p_{j|i} = \frac{w_{j|i}}{\sum_{j \neq i} w_{j|i}}
\end{equation}

The term $p_{j|i}$ captures the relative importance of the neighboring pixel $x_j$ with respect to the center pixel $x_i$. The computational complexity of this neighborhood encoding process is $O(n \times k^2)$, where $k$ denotes the kernel size, and the required space is proportional to the size of the input image.

\vspace{0.3cm}

\subsubsection{\textbf{Frequency-aware functions}}

\begin{figure}[t]
    \centering
    \includegraphics[width=0.92\linewidth]{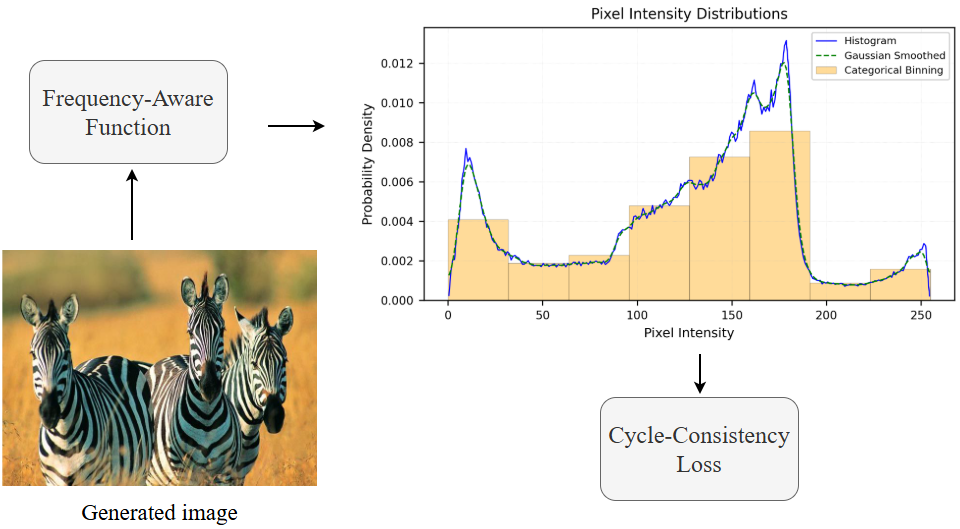}
    \caption{Visualisation of frequency-aware distribution functions applied to a generated image: (i) Histogram, (ii) Gaussian, (iii) Categorical distribution with fixed bins. These distributions are used to convert the generated image into a frequency-based representation, which is later compared using a cycle consistency loss to guide the translation network.}
    \label{fig:frequency_aware_functions}
\end{figure}

To evaluate the spectral similarity between generated and real images, we transform images into frequency-based representations using five distinct statistical distribution functions. Below is a detailed description of each method.

\begin{enumerate}
    \item \textbf{Gaussian Distribution}: models the local pixel neighborhood using a normal distribution $\mathcal{X} \sim \mathcal{N}(\mu, \sigma^2)$. The probability density function is defined as:
    \begin{equation}
        f(x|\mu, \sigma^2) = \frac{1}{\sigma\sqrt{2\pi}} e^{-\frac{1}{2} \left(\frac{x - \mu}{\sigma}\right)^2}
    \end{equation}
    For each pixel in an image of size $[256 \times 256 \times 3]$, the Gaussian distribution is computed over its local neighborhood using a kernel of size $[3 \times 3]$ (1-hop) or $[5 \times 5]$ (2-hop). This results in a computational complexity of $\mathcal{O}(H \times W \times k^2)$, where $k$ is the kernel size.

    \item \textbf{Histogram Distribution}: divides pixel intensities within a defined neighborhood into discrete bins. A kernel of size $[3 \times 3]$ or $[5 \times 5]$ slides over the image, and the histogram is computed over the corresponding pixel values. The divergence between histograms of generated and real images is then computed using KL or JS divergence. Normalization is applied before comparison. The complexity of this method is $\mathcal{O}(H \times W \times k^2 \times \log(k^2))$.

    \item \textbf{Weighted Histogram Distribution} extends the histogram approach by assigning greater weight to the central pixel of the neighborhood, with weights decreasing as distance from the center increases. This emphasizes the local structure more precisely. The complexity increases due to weighting and additional logarithmic computation, amounting to $\mathcal{O}(H \times W \times k^2 \times \log(k^2) \times \log(k))$.

    \item \textbf{Categorical Distribution} captures the frequency of distinct intensity values (or bins) for each image. The probability of each pixel category is calculated as:
    \begin{equation}
        f(x) = \frac{X_i}{\sum_i X_i}
    \end{equation}
    where $X_i$ denotes the count of intensity $i$. This method has a linear complexity of $\mathcal{O}(H \times W \times 3)$, as it operates channel-wise without overlapping neighborhoods.

    \item \textbf{Patch-wise Categorical Distribution} divides the image into non-overlapping patches of size $[8 \times 8]$. A categorical distribution is computed independently for each patch. Unlike the overlapping neighborhood-based methods (i.e., methods 1–3), this approach processes each patch discretely. The complexity is given by:
    \[
    \mathcal{O}(\text{numOfPatches} \times k^2 \times (1 + \log(k^2)))
    \]
    where $k$ is the patch size and $\text{numOfPatches}$ is $\frac{H \times W}{k^2}$.
\end{enumerate}

\begin{table*}[!htb]
    \caption{Experiment Details}
    \begin{minipage}{.45\linewidth}
    \centering
   =
    {\scriptsize
    \begin{tabular}{l l}
        \hline
         \textbf{Type} & \textbf{Description} \\
         \hline
         & 1 : Gaussian Distribution \\
         & 2 : Histogram \\
         Frequency aware functions & 3 : Weighted Histogram \\
         & 4 : Categorical distribution \\
         & 5 : Patch Categorical distribution \\
         \hline
         Loss & log : Log loss \\
         & jsd/kld: Divergence (JS/KL)\\
         \hline
         Kernel & 3 : [3,3] neighborhood \\
         & 5 : [5,5] neighborhood\\
        \hline
        cycleloss & 1 : Original Cycle Loss Used \\
                & 0 : Original Cycle Loss not used \\
        \hline
        advloss & 1 : Original Adversarial Loss used \\
        & 0 : Original Adversarial Loss not used \\
        \hline
        patchsize & [8,8]\\
        \hline
        coeff & relative importance of different objectives\\
        \hline
         
    \end{tabular}
    \label{tab:exp_parameters}
    }
    \end{minipage}%
    \begin{minipage}{.45\linewidth}
    \centering

    {\scriptsize
    \begin{tabular}{p{1cm} p{2.7cm} p{3.7cm}}
        \hline
        \textbf{Experiment}   & \textbf{Frequency-Aware}& \textbf{Model with } \\
        \textbf{name } &  \textbf{functions (ID)} & \textbf{Cycle-consistency loss} \\
        \hline
        Or                & None   & CycleGAN \\
        Or\_wt            & None   & Fd-CycleGAN  \\
        1,log             & Gaussian (1)    & Fd-CycleGAN with Log loss \\
        1,4               & Gaussian(1), \newline Categorical(4)    & Fd-CycleGAN with KLD or JSD  \\
        1,4,log           & Gaussian(1), \newline Categorical(4)    & Fd-CycleGAN withLog loss \\
        1,4,L1            & Gaussian(1), \newline Categorical(4)    & Fd-CycleGAN with L1 \\
        2,4               & Hisogram(2), \newline Categorical(4)    & Fd-CycleGAN with KLD or JSD  \\
        3,4               & Weighted Histogram (3), \newline Categorical(4)    & Fd-CycleGAN with KLD or JSD  \\
        log               & None    & Fd-CycleGAN with Log loss \\
        log\_k3\_wt       & None      & Fd-CycleGAN with Log loss on weighted image (kernel = 3) \\
        log\_k5\_wt       & None     & Fd-CycleGAN with Log loss (kernel = 5) \\
        kld               & None   & Fd-CycleGAN with KL Divergence \\
        jsd               & None   & Fd-CycleGAN with JS Divergence \\
        \hline
    \end{tabular}
    }
    \label{tab:exp_details}
    \end{minipage}
\end{table*}

\subsection{Translation Loss Functions}
In our proposed model, the original cycle-consistency loss is replaced with a divergence-based formulation, while the adversarial and identity losses are retained \cite{zhu2017_CycleGAN}. The model learns bidirectional mappings $\mathcal{T}$ and $\mathcal{F}$ using a combination of adversarial, cycle-consistency, and identity losses. The overall objective function is defined as:

\begin{equation} \label{eq:adversarial_loss}
\begin{split}
    \mathcal{L}(\mathcal{G}, \mathcal{D}) := &\ \mathcal{L}_{adv}(\mathcal{G}_1, \mathcal{G}_2, \mathcal{D}_1, \mathcal{D}_2) + \lambda_{cyc} \cdot \mathcal{L}_{cyc}(\mathcal{G}_1, \mathcal{G}_2) \\
    &+ \lambda_{id} \cdot \mathcal{L}_{id}(\mathcal{G}_1, \mathcal{G}_2)
\end{split}
\end{equation}

Here, $\mathcal{G}_1$, $\mathcal{G}_2$ are the generators, and $\mathcal{D}_1$, $\mathcal{D}_2$ are the corresponding discriminators for domains $\mathcal{X} \in \mathcal{P}$ and $\mathcal{Y} \in \mathcal{Q}$. In our proposed model, the original cycle-consistency loss is replaced with a divergence-based formulation, while the adversarial and identity losses are retained.

\subsubsection{\textbf{Adversarial Loss}}

The adversarial component encourages the generators to produce outputs to be indistinguishable from real samples. We employ the logarithmic loss formulation for the discriminators and generators. The adversarial loss is given by:

\begin{equation}
\begin{split}
    \mathcal{L}_{adv}(\mathcal{G}_1, \mathcal{G}_2, \mathcal{D}_1, \mathcal{D}_2) &= \mathbb{E}_{x,y}[\log \mathcal{D}(x, y)] \\
    &+ \mathbb{E}_{x,y}[1 - \log \mathcal{D}(\mathcal{G}(x, y))]
\end{split}
\end{equation}

\subsubsection{\textbf{Cycle-consistency Loss}} ensures that mapping an image to the target domain and back reconstructs the original image. In this work we explore three formulations:

\begin{itemize}
    \item \textbf{L1 Norm}:
    The original CycleGAN uses the pixel-wise L1 norm to enforce cycle-consistency:
    \begin{equation}
    \begin{split}
        \mathcal{L}_{cyc}^{L1}(\mathcal{G}_1, \mathcal{G}_2) &= \mathbb{E}_{x \sim p_{data}}[\| \mathcal{G}_2(\mathcal{G}_1(x)) - x \|_1] \\
        &+ \mathbb{E}_{y \sim p_{data}}[\| \mathcal{G}_1(\mathcal{G}_2(y)) - y \|_1]
    \end{split}
    \end{equation}
    While computationally efficient $\mathcal{O}(H \times W)$, it fails to capture local spatial patterns and neighborhood-level statistics.

    \item \textbf{KL Divergence and JS Divergence}:
    To better model local distributions, we replace the L1 norm with statistical divergences. The KL divergence between distributions $\mathcal{P}$ and $\mathcal{Q}$ is:
    \begin{equation}
        KL(\mathcal{P} \| \mathcal{Q}) = \sum_i \sum_j p_{ij} \log \frac{p_{ij}}{q_{ij}}
    \end{equation}
    Note that $KL(\mathcal{P} \| \mathcal{Q}) \ne KL(\mathcal{Q} \| \mathcal{P})$, indicating its asymmetry. To address asymmetry and stability, we also consider the symmetric Jensen-Shannon (JS) divergence:
    \begin{equation}
    \begin{split}
        JSD(\mathcal{P} \| \mathcal{Q}) &= \frac{1}{2} KL\left(\mathcal{P} \| \frac{\mathcal{P} + \mathcal{Q}}{2}\right) \\
        &+ \frac{1}{2} KL\left(\mathcal{Q} \| \frac{\mathcal{P} + \mathcal{Q}}{2}\right)
    \end{split}
    \end{equation}
    JS divergence yields a bounded and symmetric loss:
    \begin{equation}
        0 \leq JSD(\mathcal{P} \| \mathcal{Q}) \leq 1
    \end{equation}
    The overall divergence-based cycle-consistency loss becomes:
    \begin{equation}
    \begin{split}
        \mathcal{L}_{cyc}^{div} = StD(\mathcal{X}_{real} \| \mathcal{X}_{rec}) + StD(\mathcal{Y}_{real} \| \mathcal{Y}_{rec})
    \end{split}
    \end{equation}
    This formulation captures the structural and statistical alignment of generated and real domains with complexity $\mathcal{O}(H \times W \times \log(H \times W))$ with statistical divergence, $StD$ (KLD or JSD).

    \item \textbf{Log Loss}: we also experiment with log loss as a reguloarizer to penalize deviation between real and reconstructed distributions:
    \begin{equation}
    \begin{split}
        \mathcal{L}_{log}(\mathcal{G}) = \mathbb{E}_{X}[-X_{real} \log(X_{rec}) \\
        - (1 - X_{real}) \log(1 - X_{rec})]
    \end{split}
    \end{equation}
   
    This loss maintains low computational complexity ($\mathcal{O}(H \times W)$), while providing sharper convergence and better regularization.
\end{itemize}

All three loss formulations aim to improve unsupervised domain translation by capturing either pixel-wise fidelity (L1), statistical similarity (KL/JS), or probabilistic regularization (log loss). Their computational cost remains manageable for fixed image sizes (e.g., $256 \times 256 \times 3$).

\subsubsection{\textbf{Identity Loss}} is used to preserve color composition and content during translation. It encourages generators to behave like identity mappings when fed samples from the target domain:

\begin{equation}
\begin{split}
    \mathcal{L}_{id} &= \mathbb{E}_{y \sim p_{data}}[\| \mathcal{G}_2(y) - y \|_1] \\
    &+ \mathbb{E}_{x \sim p_{data}}[\| \mathcal{G}_1(x) - x \|_1]
\end{split}
\end{equation}

While this loss improves consistency and preserves fine-grained structure, it may limit the generative diversity if over-penalized. Hence, its weight $\lambda_{id}$ is tuned based on task-specific requirements.

By incorporating multiple statistical representations, ranging from Gaussian and histogram-based measures to categorical distributions. We enable the model to capture local spatial and spectral nuances in the image data. These structured enhancements allow the model to preserve semantic content while effectively translating images with diverse strike-off patterns, ensuring both visual fidelity and contextual integrity in the translated outputs.

\section{Experiments}\label{sec:exp}

In this section, we systematically investigate the impact of incorporating frequency distribution functions and divergence-based and logarithmic loss formulations into the CycleGAN framework. Our primary goal is to assess how these enhancements influence the quality and fidelity of I2I translation. We conduct a series of experiments to compare the performance of the proposed Fd-CycleGAN with the baselines, like CycleGAN, Diffusion models and CUT model. A comprehensive ablation study is presented to evaluate the individual contributions of each frequency-aware functions. To ensure a thorough evaluation, we employ widely used quantitative metrics, including Peak Signal-to-Noise Ratio (PSNR), Structural Similarity Index (SSIM), F1 Score, Fréchet Inception Distance (FID), and CLIP score, thereby validating the effectiveness of our proposed approach across various settings and datasets.

\begin{table*}[]
    \centering
    \caption{A quantitative comparison of different Experiments for Horse to Zebra and Zebra to Horse at object level $(256 \times 256)$}
    \begin{tabular}{l l l l l l l l l l l l l }
    \hline
         \textbf{Task} & & \multicolumn{5}{c}{Horse to Zebra} & & \multicolumn{5}{c}{Zebra to Horse}\\
	\hline
\textbf{Metrics}		&	&	\textbf{PSNR}$\uparrow $	&	\textbf{SSIM}$\uparrow $	&	\textbf{f1}$\uparrow $	&$\uparrow $	\textbf{FID}$\downarrow$	&	\textbf{CLIP}$\uparrow $	&	&	\textbf{PSNR}$\uparrow $	&	\textbf{SSIM}$\uparrow $	&	\textbf{f1}$\uparrow $	&	\textbf{FID}$\downarrow$	&	\textbf{CLIP}$\uparrow $	\\ \hline
CycleGAN 		&	&	23.629	&	0.819	&	0.768	&	67.822	&	28.087	&	
&	23.566	&	0.836	&	0.754	&	179.557	&	23.395	\\
CycleGAN with \textit{wt\_image}		&	&	25.987	&	0.890	&	0.773	&	81.675	&	29.131	&	
&	24.181	&	0.881	&	0.797	&	170.382	&	24.949	\\
Gauss (log)		&	&	14.042	&	0.393	&	0.580	&	73.674	&	25.856	&	
&	13.292	&	0.324	&	0.591	&	142.168	&	26.472	\\
Gauss+Catg		&	&	23.051	&	0.844	&	0.691	&	81.950	&	29.017	&	
&	22.418	&	0.859	&	0.706	&	174.967	&	25.858	\\
Gauss+Catg (log) 		&	&	23.076	&	0.842	&	0.762	&	78.250	&	21.309	&	
&	22.669	&	0.853	&	0.759	&	165.433	&	76.879	\\
Gauss+Catg (L1) 		&	&	25.907	&	0.883	&	0.780	&	72.822	&	28.911	&	
&	25.811	&	0.902	&	0.779	&	165.626	&	23.918	\\
Gauss+Catg on\textit{ wt\_image	}	&	&	24.599	&	0.885	&	0.789	&	74.734	&	21.628	&	
&	13.717	&	0.336	&	0.587	&	156.430	&	77.711	\\
Hist+Catg		&	&	23.814	&	0.843	&	0.742	&	78.407	&	28.100	&	
&	22.906	&	0.862	&	0.714	&	158.507	&	25.977	\\
Hist+Catg on \textit{wt\_image}		&	&	24.332	&	0.880	&	0.793	&	82.333	&	28.133	&	
&	23.403	&	0.889	&	0.777	&	165.411	&	24.769	\\
wtHist+Catg on \textit{wt\_image}		&	&	21.759	&	0.830	&	0.717	&	78.518	&	28.196	&	
&	22.518	&	0.854	&	0.738	&	166.246	&	24.997	\\
Log on \textit{wt\_image}		&	&	21.066	&	0.533	&	0.585	&	69.874	&	24.119	&	
&	19.906	&	0.332	&	0.608	&	172.146	&	22.358	\\
kld on \textit{wt\_image}		&	&	24.241	&	0.870	&	0.778	&	76.286	&	22.238	&	
&	23.455	&	0.882	&	0.767	&	167.232	&	21.578	\\
jsd		&	&	22.404	&	0.781	&	0.774	&	77.063	&	28.366	&	
&	22.265	&	0.803	&	0.754	&	184.783	&	22.141	\\
jsd on \textit{wt\_image}		&	&	24.548	&	0.863	&	0.765	&	67.535	&	22.111	&	
&	24.531	&	0.880	&	0.769	&	163.954	&	22.042	\\
         \hline
         \multicolumn{13}{l}{} \\
         \multicolumn{13}{c}{\scriptsize{The experiments have been carried out with different combinations of various FDs: 1) Gaussian distribution (Gauss) 2) Histogram distribution (Hist) }}\\
         \multicolumn{13}{c}{\scriptsize{ 3) weighted Histogram distribution (WtHist) 4) Categorical distribution (Catg) on original or weighted images (wt\_image) with different loss }}\\
         \multicolumn{13}{c}{\scriptsize{ functions as  Log, KLD, JSD losses. In specific cases where losses are computed along with FD computation is mentioned in the brackets.}}\\
    \end{tabular}
    
    \label{tab:metrics}
\end{table*}

\subsection{Experimental Setup}

\textbf{Dataset Details}.
We conducted our experiments using three datasets: Horse2Zebra, Monet2Photo, and the Strike-off dataset. The first two datasets, introduced in \cite{zhu2017_CycleGAN}, are standard benchmarks for style transfer and I2I translation. The Strike-off dataset, proposed in \cite{nigam2022deformity} and also used in document analysis and recognition tasks \cite{surveyDAR2023, nigam2023strike}, presents a real-world challenge of removing noisy strike-off patterns from handwritten documents. These datasets vary significantly in terms of color spectra, texture distributions, and domain characteristics. All images were resized to $256 \times 256 \times 3$ and normalized before training. Dataset splits for training and testing are detailed in Table \ref{tab:data set}.

\textbf{Implementation Details.} 
All experiments were conducted on an NVIDIA RTX 3090 GPU with 24GB VRAM, using PyTorch 1.13. The models were trained using the Adam optimizer with learning rate 0.0002, batch size 1, and $\beta_1 = 0.5$ for 200 epochs. For evaluation, we used the following metrics: PSNR, SSIM, FID, CLIPScore, and F1 (discussed further). Full experimental configurations are provided in Table \ref{tab:exp_details}. Our experiments are categorized into two settings: 1) Fd-CycleGAN on original images, and 2) Fd-CycleGAN on weighted images, where weighted images are generated using local neighborhood encoding to emphasize spatial relevance. The model architecture closely follows the original CycleGAN design \cite{zhu2017_CycleGAN}. The generator employs ResNet blocks integrated with convolutional and deconvolutional layers, while the discriminator uses a $70 \times 70$ PatchGAN. 

\textbf{Evaluation Metrics.} To evaluate translation consistency and image quality, we employed a combination of standard metrics:
1) \textit{PSNR} (Peak Signal-to-Noise Ratio): Measures signal fidelity; higher values indicate greater similarity to the target. 2) \textit{SSIM} (Structural Similarity Index): Evaluates structural content such as luminance, contrast, and texture between the generated and reference images. 3) \textit{F1 Score}: Assesses a balance between precision and recall, particularly useful in document contexts for binary classification (text vs. strike-off). 4) \textit{FID} (Fréchet Inception Distance): Quantifies the difference in feature distributions between real and generated images using a pre-trained Inception network. Lower scores imply better visual quality. 5) \textit{CLIP Score}: Measures semantic alignment between images and textual concepts, indicating how well the generated image captures the intended semantics. Higher values denote better alignment.

\textbf{Baselines.} We benchmark our proposed model against the original CycleGAN \cite{zhu2017_CycleGAN}, CUT \cite{CUT2020contrastive}, Diffusion models CycleDiff\cite{Diffusion_1_2020denoising, diffusion_2rombach2022high} to highlight improvements in both visual quality and quantitative metrics.

\begin{figure*}
    \centering
    \includegraphics[scale=0.265]{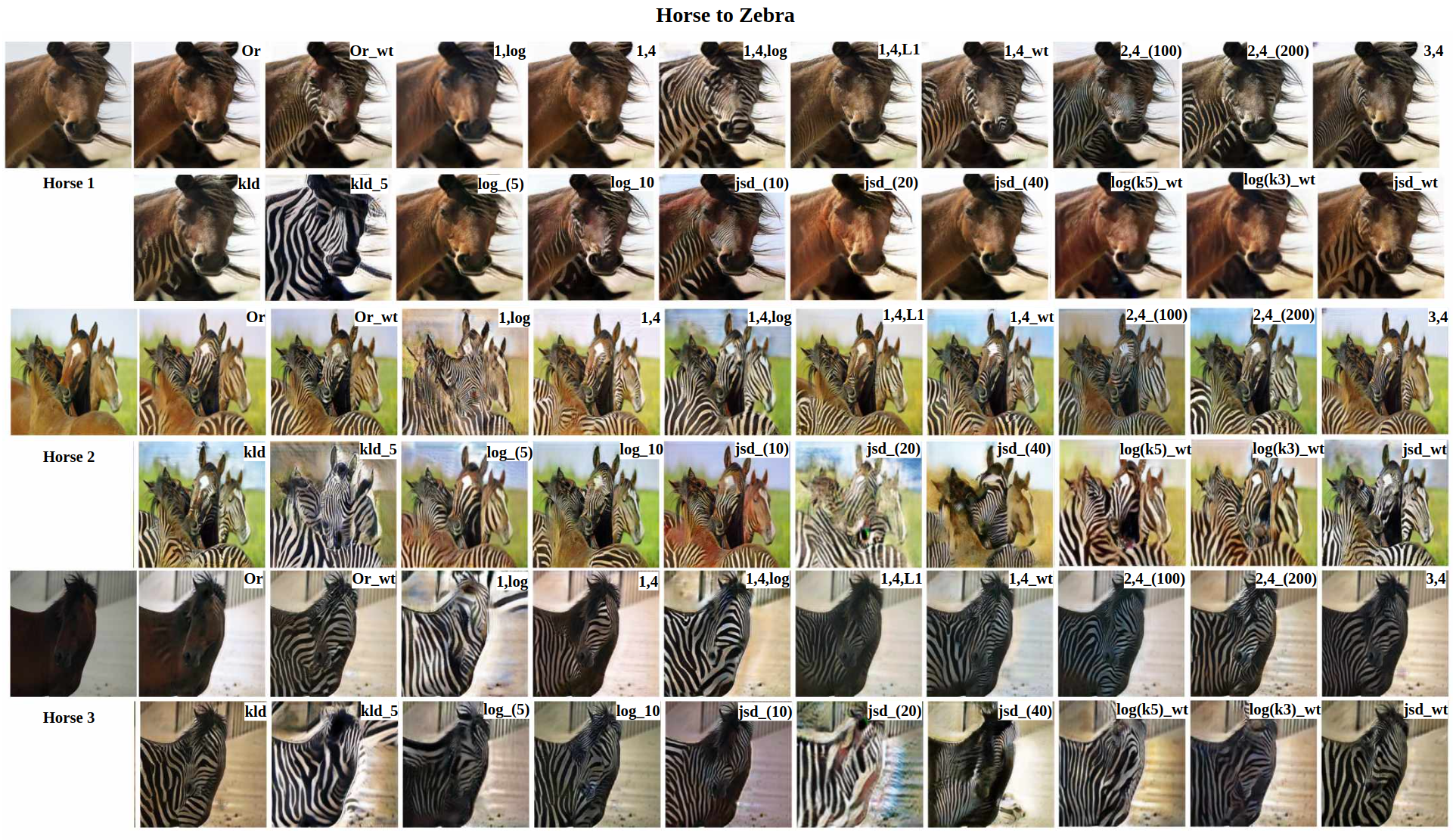}
    \caption{Latent representations learning on Horse2Zebra data set for Horse to Zebra translation with various combinations of frequency distributions and loss metrics. Frequency distributions: 1- Gaussian distribution, 2-Histogram 3-Weighted histogram 4-Categorical distribution 5-Categorical patch distribution. Loss metrics: Divergence (jsd, kld), log}
    \label{fig:h2z_wt_cyclegan_all}
\end{figure*}

\begin{figure*}
    \centering
    \includegraphics[scale=0.4]{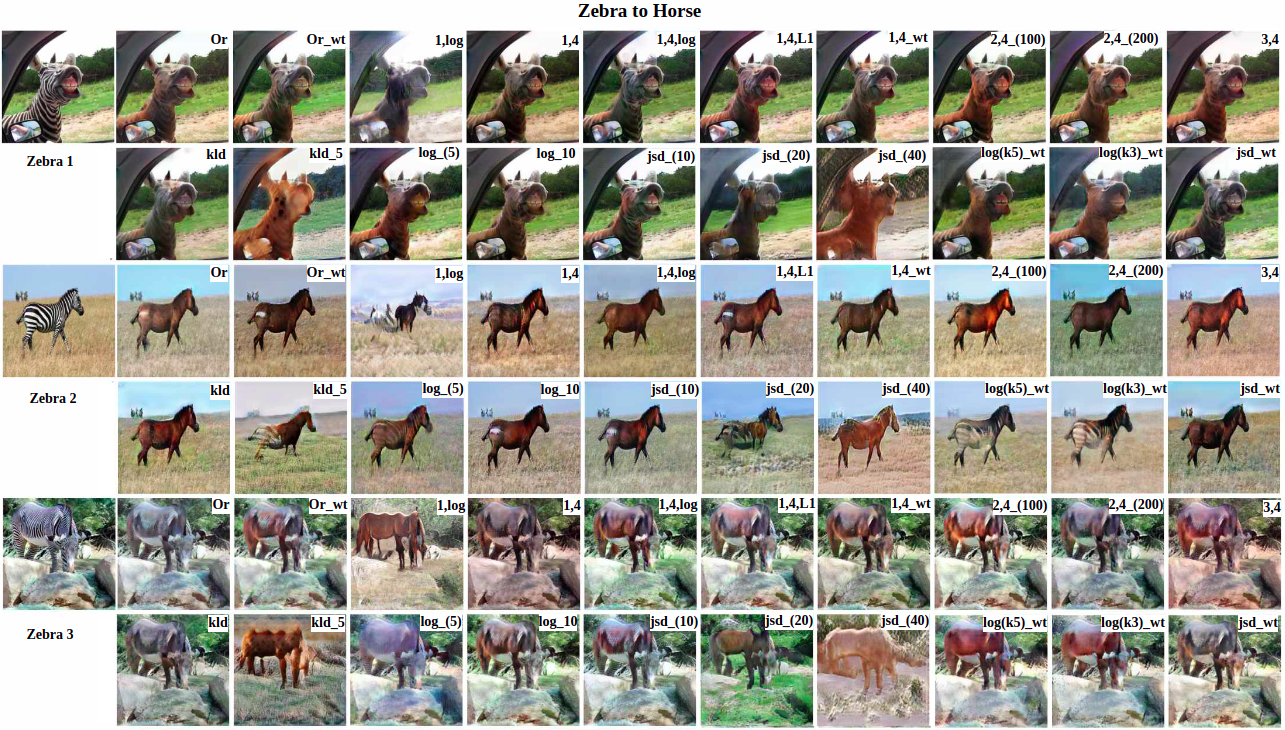}
    \caption{Latent representations learning on Zebra2Horse data set for Zebra to Horse translation with various combinations of frequency distributions and loss metrics. Frequency distributions: 1- Gaussian distribution, 2-Histogram 3-Weighted histogram 4-Categoricalelaborate distribution 5-Categorical patch distribution. Loss metrics: Divergence (jsd, kld), log}
    \label{fig:z2h_wt_cyclegan_all}
\end{figure*}

\begin{figure}[t]
    \centering
    \includegraphics[scale=0.17]{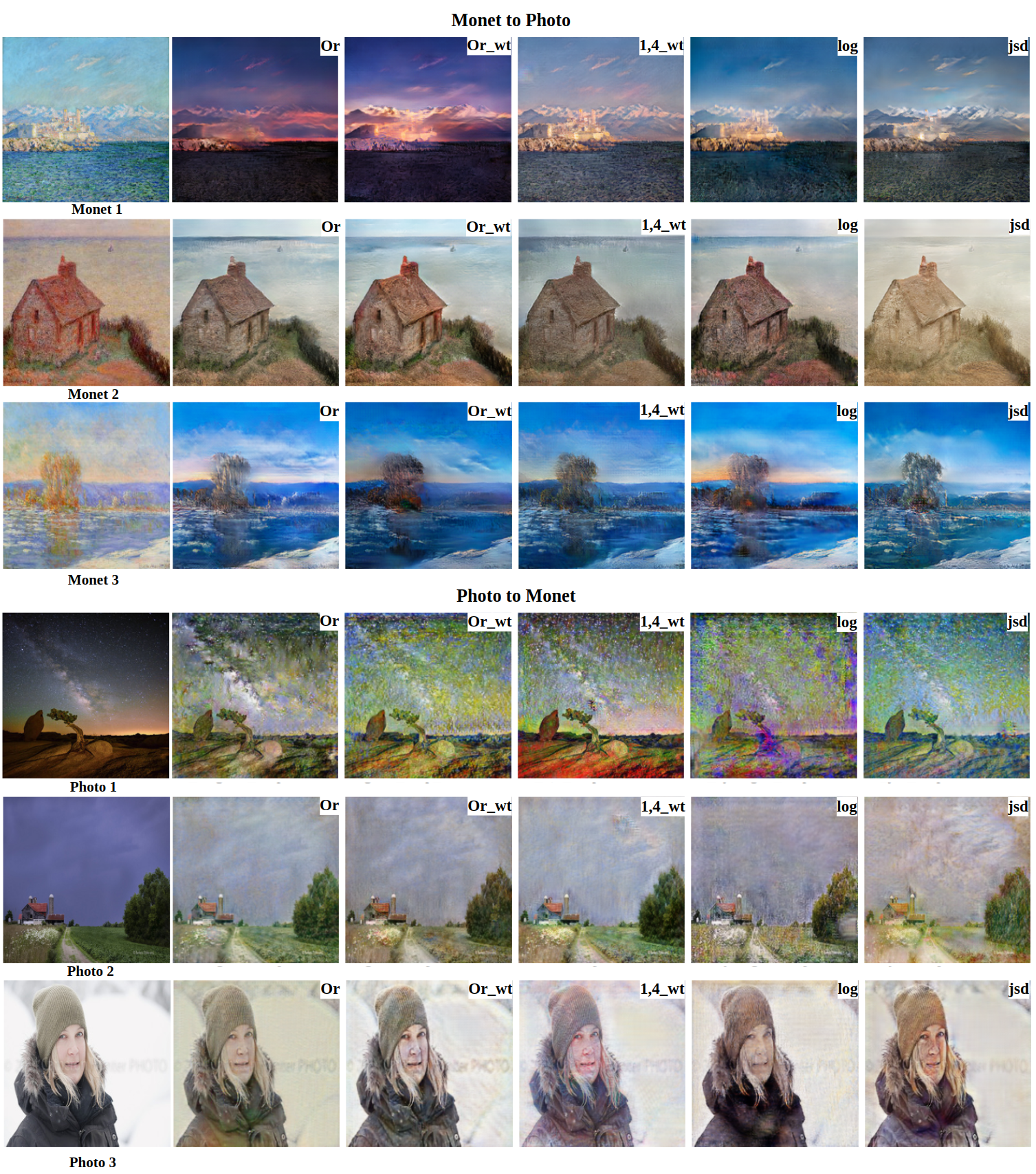}
    \caption{Results of Latent representations learning on Monet2Photo data set for Monet to Photo and vice-versa translation with various combinations of frequency distributions and loss metrics. Frequency distributions: 1- Gaussian distribution, 2-Histogram 3-Weighted histogram 4-Categorical distribution 5-Categorical patch distribution. Loss metrics: Divergence (jsd, kld), log}
    \label{fig:m2p_cyclegan_all}
\end{figure}

\begin{figure}[t]
    \centering
    \includegraphics[scale=0.17]{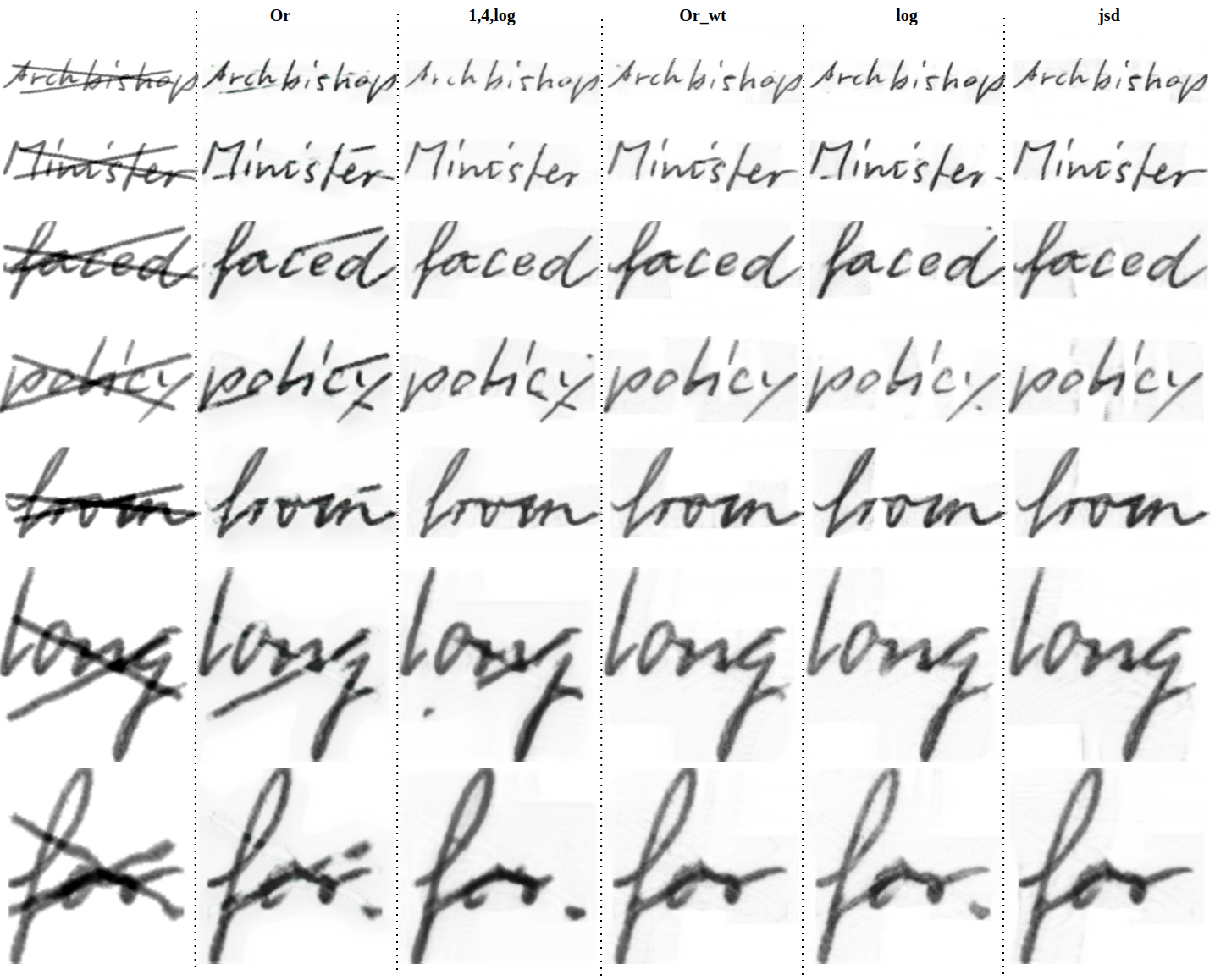}
    \caption{Results of Latent representations learning on Strike-off data set for Strike-off removal with various combinations of frequency distributions and loss metrics. Frequency distributions: 1- Gaussian distribution, 2-Histogram 3-Weighted histogram 4-Categorical distribution 5-Categorical patch distribution. Loss metrics: Divergence (jsd, kld), log}
    \label{fig:strikeoff_all}
\end{figure}

\begin{table}[]
    \centering
    \caption{Data set details}
    {\scriptsize
    \begin{tabular}{l l l l l }
    \hline
    \textbf{Data set} & \textbf{trainA} & \textbf{trainB} & \textbf{testA} & \textbf{testB}\\
    \hline
    Horse2Zebra (H2Z) & 1067 & 1334 & 120 & 140 \\
    Monet2Photo (M2P)& 1072 & 6287 & 121 & 751 \\
    Strike-off (Str) & 402 & 402 & 34 & 34\\
    \hline
    \end{tabular}
    }
    \label{tab:data set}
\end{table}

\begin{figure*}[ht]
    \centering
    \includegraphics[scale=0.18]{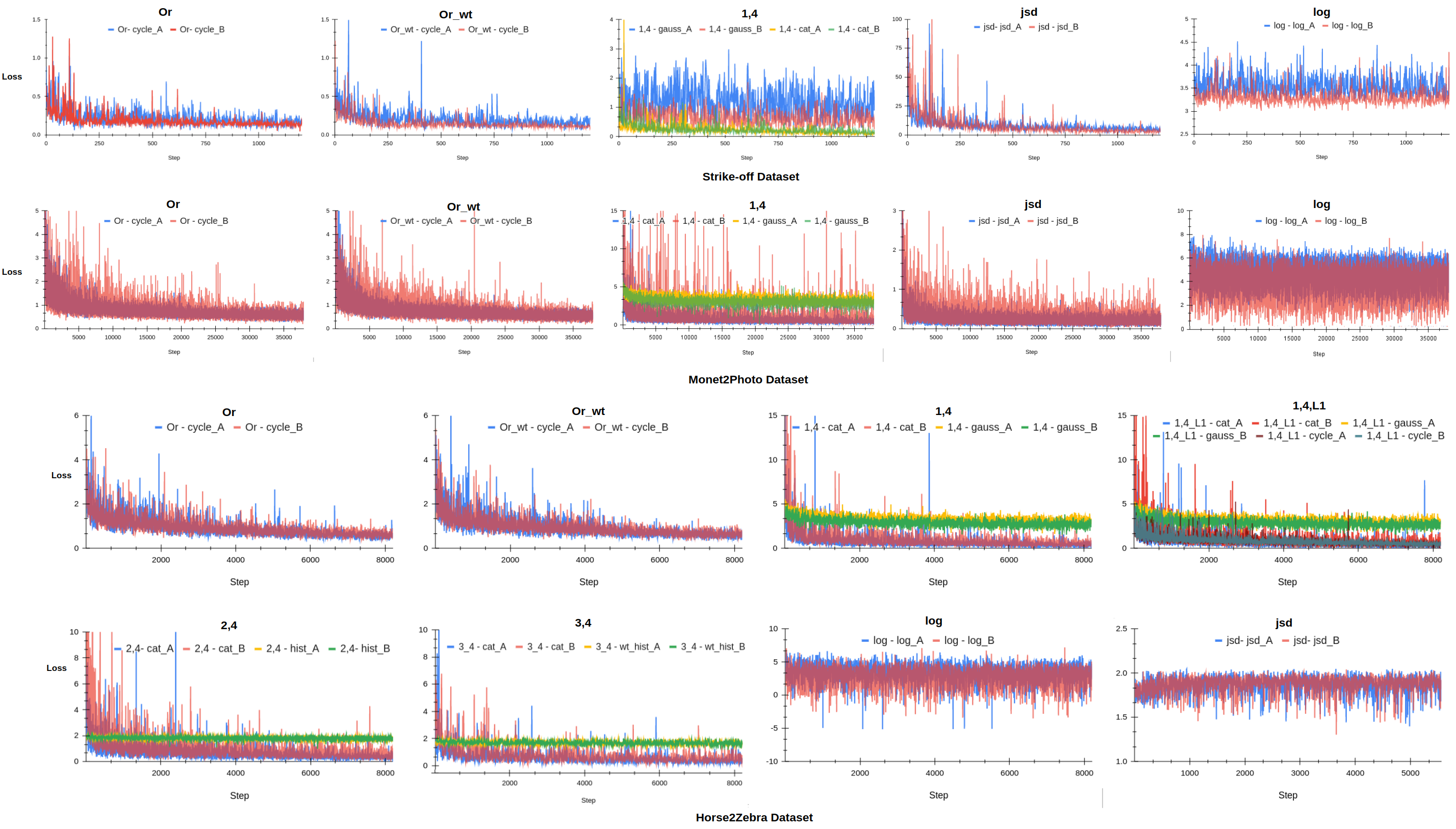}
    \caption{Performance graphs of various frequency-aware functions (1,2,3,4) and losses (jsd, kld, log) on Horse2Zebra, Monet2Photo and Strike-off removal data sets. }
    \label{fig:graph_all}
\end{figure*}

\subsection{Ablation Study}
To analyze the contribution of each component, we systematically varied the type of frequency-aware functions (FD-functions), loss functions, and image weighting (wt\_image). Table \ref{tab:metrics} of evaluation metric results shows that using categorical distributions with histogram-based FD improves FID and SSIM scores compared to Gaussian or log-loss variants. Notably, the JSD loss on weighted images yields a strong balance between PSNR (24.54) and FID (67.53). We present a detailed comparison of various ablation experiments conducted on multiple datasets in Figures \ref{fig:h2z_wt_cyclegan_all}, \ref{fig:z2h_wt_cyclegan_all}, \ref{fig:m2p_cyclegan_all}, and \ref{fig:strikeoff_all}. Figure \ref{fig:graph_all} illustrates the comparative performance of different Frequency Distribution (FD) functions employed for the I2I translation task. These FDs facilitate the learning of local structures within the images, thereby guiding the generator to synthesize outputs that align more closely with the underlying data distributions. To further enhance the semantic fidelity of the translated images, we proposed a Local Neighborhood Encoding (LNE) to preserves the contextual integrity of the local regions. It improves the visual realism and texture quality of the generated images significantly. Our experiments demonstrate that the effectiveness of each FD combination varies, with certain configurations producing superior translation results.

Notably, the configurations \textit{L1 on wt\_image}, \textit{Gauss+Catg (log loss)}, \textit{Gauss+Catg (L1 loss)}, \textit{Gauss+Catg}, \textit{Hist+Catg}, \textit{log loss on wt\_image}, and \textit{jsd on wt\_image} have consistently outperformed other variants. Among them, \textit{L1 on wt\_image} yields images with perceptually higher similarity to the target domain, demonstrating sharper edges and more vibrant color contrasts compared to the baseline CycleGAN (\textit{L1}). The \textit{KLD} and \textit{JSD} variants exhibit visually similar outputs, indicating their potential interchangeability. The \textit{kld\_patch} experiment successfully captures the semantic characteristics of the target domain; however, the use of patch-based learning introduces a slight blurriness, suggesting the necessity of pixel-level learning for sharper outputs. Experiments employing logarithmic loss, such as \textit{log}, $\textit{log with k=3 on wt\_image}$, and $\textit{log with k=5 on wt\_image}$—achieved successful style transfer. Nevertheless, the outputs display low contrast and mild blur. Interestingly, the combination of categorical distribution with log loss demonstrates superior performance overall, while Gaussian and Histogram distributions combined with categorical information deliver even better translation quality.

Quantitative results are summarized in Table \ref{tab:metrics}. Experiments involving '\textit{Gaussian}', '\textit{Histogram}', and '\textit{weighted images}' consistently achieve higher PSNR, SSIM, F1, CLIP scores, indicating better fidelity and perceptual quality. The use of Gaussian distribution in conjunction with categorical information proves especially effective in capturing semantic structures critical for successful translation. Additionally, weighted versions of the experiments surpass their non-weighted counterparts, emphasizing the benefits of incorporating region-specific importance in translation. While most experiments result in strong FID scores, the 'log'-based variants perform comparatively lower, suggesting less realism. Moreover, the ‘weighted Histogram’ variant outperforms the basic 'Histogram', reaffirming that incorporating weight improves translation quality. Overall, all frequency-based methods enhance both image realism and translation consistency, with weighing mechanisms further contributing to perceptual improvements. Despite a modest increase in computational cost, Fd-CycleGAN demonstrates clear advantages by generating higher-quality images with finer textures and better semantic alignment.



\subsection{Computational Cost Analysis}
\begin{figure*}
    \centering
    \includegraphics[width=0.99\linewidth]{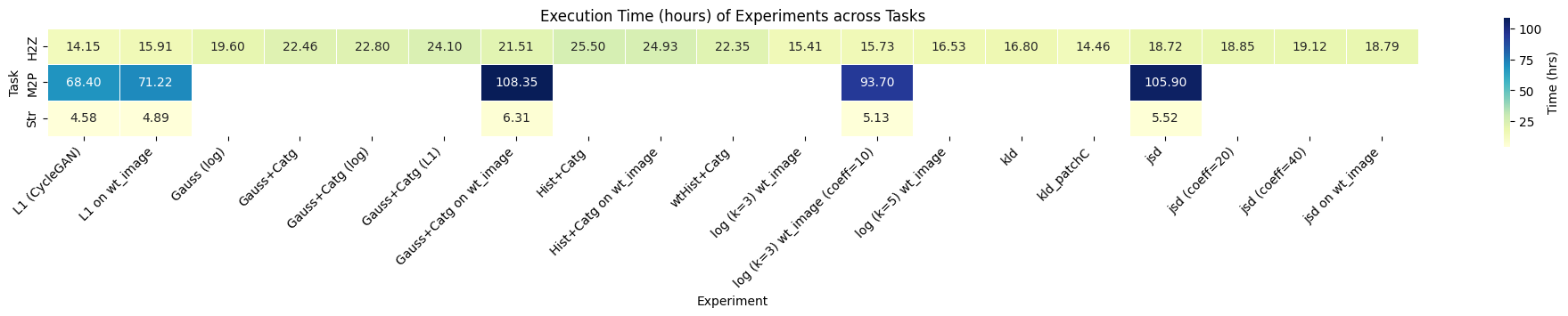}
    \caption{Computational cost of different Experiments for various coefficients pairs (Total Runtime in hours). The coefficients regulate the impact of different objective functions on total adversarial loss}
    \label{fig:heatmap}
\end{figure*}

We analyzed per-experiment runtimes and present a heatmap in Figure \ref{fig:heatmap}. Models using log loss on weighted images showed faster convergence, whereas divergence-based models were computationally heavier but yielded better FID. Histogram-based models demonstrated a good balance of speed and quality. The heatmap clearly shows a wide variation in training times across tasks and loss functions. M2P (Monet2Photo) experiments, especially with composite losses like Gauss+CatG, Hist+CatG, and jsd, show very high training time (up to 108.35 hours), indicating computational complexity. In contrast, the Str (strike-off) task shows much lower training time, suggesting either simpler data or faster convergence. Consistent trends are seen across tasks where divergence-based losses (jsd, log, Hist+CatG) take significantly more time compared to basic L1-based CycleGAN. The heatmap supports the claim that domain-specific complexity affects convergence, reinforcing your proposed domain-sensitive architecture. There's an observable trade-off between performance (as described in your metric tables) and execution time. Models with log, jsd, or Hist+CatG losses perform better but require more computational resources. This supports the argument for evaluating training time vs accuracy trade-offs. The training time for M2P using Hist+CatG reaches 108.35 hours, while Str completes within 6.31 hours using the same configuration. This stark contrast demonstrates the domain-dependent convergence behavior. Despite the increased complexity, divergence-based losses significantly enhanced the output quality for both natural and artistic domains.

\begin{table}
    \centering
    \caption{Comparison of different Experiments for Horse to Zebra at object level $(256 \times 256)$}
    \begin{tabular}{l l l l l }
    \hline
         \textbf{Task} & \multicolumn{4}{c}{Horse to Zebra} \\
	\hline
\textbf{Metrics}		
&	\textbf{PSNR}$\uparrow $	
&	\textbf{SSIM}$\uparrow $	
&	\textbf{FID}$\downarrow$	
&	\textbf{CLIP}$\uparrow $		\\ \hline
CycleGAN
&		18.53
&		0.67
&		77.18
&		28.07	\\ 
CUT	\cite{CUT2020contrastive}
&		13.71
&		0.35
&		45.50
&		29.15	\\ 
CycleDiff	\cite{Diffusion_1_2020denoising}	
&		11.51
&		0.21
&		347.27
&		25.04	\\ 
CycleNet	\cite{diffusion_2rombach2022high}
&		20.42
&		0.52
&		81.69
&		28.91  \\

\hline
Fd-CycleGAN(L1)		&		25.98	&	0.89	&		81.67	&	29.13\\
Fd-CycleGAN(Gauss+Catg)			&	23.05	&	0.84	&		81.95	&	29.01\\
Fd-CycleGAN(Hist+Catg)			&	23.81	&	0.83	&	78.40	&	28.10	\\
Fd-CycleGAN(Log)		&		21.06	&	0.53	&		69.87	&	24.11	\\
Fd-CycleGAN(KLD)		&		24.24	&	0.87	&		76.28	&	22.23	\\
Fd-CycleGAN(JSD) 	&		24.54	&	0.86	&		67.53	&	22.11	\\
\hline

    \end{tabular}
    
    \label{tab:metrics1}

\end{table}

\subsection{Performance Summary}
From Table \ref{tab:metrics}, we observe that, \textit{Fd-CycleGAN with L1 loss} achieves the highest PSNR (25.98) and SSIM (0.89). \textit{JSD on weighted image} offers the best FID score (67.53), suggesting better realism. \textit{Histogram + Categorical distributions} offer a good trade-off between fidelity (PSNR: 23.81) and diversity (FID: 78.40). All reported scores are averaged over three independent runs. We conducted a Wilcoxon signed-rank test between CycleGAN and our best model, obtaining $p < 0.05$ for FID and SSIM, indicating statistical significance.

\section{Discussion}\label{sec:disc}

\subsection{Impact of Frequency-aware Functions}

In this work, we introduced frequency distribution (Fd) functions to guide the model in learning richer latent representations of the images, thereby enhancing its ability to capture the semantic structure of the target domain. Our experimental results confirm that the incorporation of frequency distribution functions significantly enhances the performance of Fd-CycleGAN. These functions allow the generator to better align the local statistical properties of the generated images with those of the target domain. Specifically, the combination of frequency-based representations and cycle-consistency loss enables the network to produce visually superior and semantically meaningful translations. Different FD functions exhibited varied effects on performance across datasets. This highlights the importance of selecting appropriate frequency distributions based on the nature of the data and the translation task. Moreover, LNE further demonstrated improved image quality, indicating that emphasizing key image regions during translation benefits the overall perceptual quality. In conclusion, the use of frequency distribution functions not only enhances convergence and translation quality but also enables the model to learn more semantically aligned and perceptually consistent mappings between source and target domains.

\vspace{0.1cm}

\textit{Gaussian Distribution Function}
The results obtained using the Gaussian distribution function are visually appealing, showcasing an enhanced color palette and improved detail preservation in the generated images, as illustrated in Figure~\ref{fig:h2z_wt_cyclegan_all} for the following categories: \textit{Gauss+Catg (log loss)}, \textit{Gauss+Catg (L1 loss)}, and \textit{Gauss+Catg}. These results indicate smoother image translation and notable improvement in fine-grained details.This enhanced performance is achieved by combining two complementary components: the local neighborhood preservation encoded by the Gaussian distribution function (\textit{Gauss}) and the global structural modeling provided by the Categorical distribution function (\textit{Catg}). While the Gaussian distribution captures local structures and subtle textures, the categorical function helps in aligning the overall semantic layout of the images. Among the combinations tested, the use of Gaussian and categorical distributions with log loss (\textit{Gauss+Catg with log loss}) produced the most structurally faithful results when compared to the ground truth. The effectiveness of the Gaussian distribution lies in its ability to model continuous and symmetric data variations, which leads to better preservation of continuous image features such as color gradients, texture consistency, and object shape. It also retains fine class-specific details and supports style diversity—such as variations in lighting and background—without introducing artifacts. Furthermore, the local structure encoding enabled by the Gaussian function plays a critical role in addressing mode collapse. By ensuring that diverse local variations in the target domain are reflected in the generated samples, it promotes more stable and generalizable training. Overall, the Gaussian distribution function significantly enhances the perceptual quality, structural accuracy, and robustness of image translation in our proposed framework.
    

\textit{Histogram and Weighted Histogram}
The Histogram and Weighted Histogram functions are effective in capturing multi-modal data distributions, as evidenced by the results in Figure~\ref{fig:h2z_wt_cyclegan_all}, particularly in categories $2$ and $3$, which correspond to combinations with the Categorical distribution function ($4$). These frequency-based representations demonstrate an ability to model the intrinsic properties of the data, enabling more effective and semantically meaningful image translation. The Histogram function, by representing the discrete intensity values of image pixels, facilitates pixel-level translation from the target domain to the generated samples. This capability proves especially beneficial in capturing and transferring diverse styles present in the target distribution, leading to visually rich and varied outputs. Compared to the Gaussian distribution, the Histogram-based methods exhibit greater robustness to outliers—an observation that is reflected in the consistency and stability of style elements such as background texture and lighting across translated images. A notable strength of these methods lies in their direct adaptation from the empirical distribution of real data, which the generator learns and mimics. This results in improved alignment between generated and real samples without requiring strong prior assumptions about the underlying data distribution. Additionally, Histogram-based approaches are more computationally efficient than Gaussian-based ones, due to the simpler nature of their transformations. The binning mechanism inherent to Histogram functions also introduces a level of flexibility, allowing control over the granularity of features captured. This makes it possible to tune the model’s sensitivity to subtle or broad patterns in the data, enhancing the adaptability of the translation process across varying domains and complexities.

\begin{figure*}
    \centering
    \includegraphics[scale=0.2]{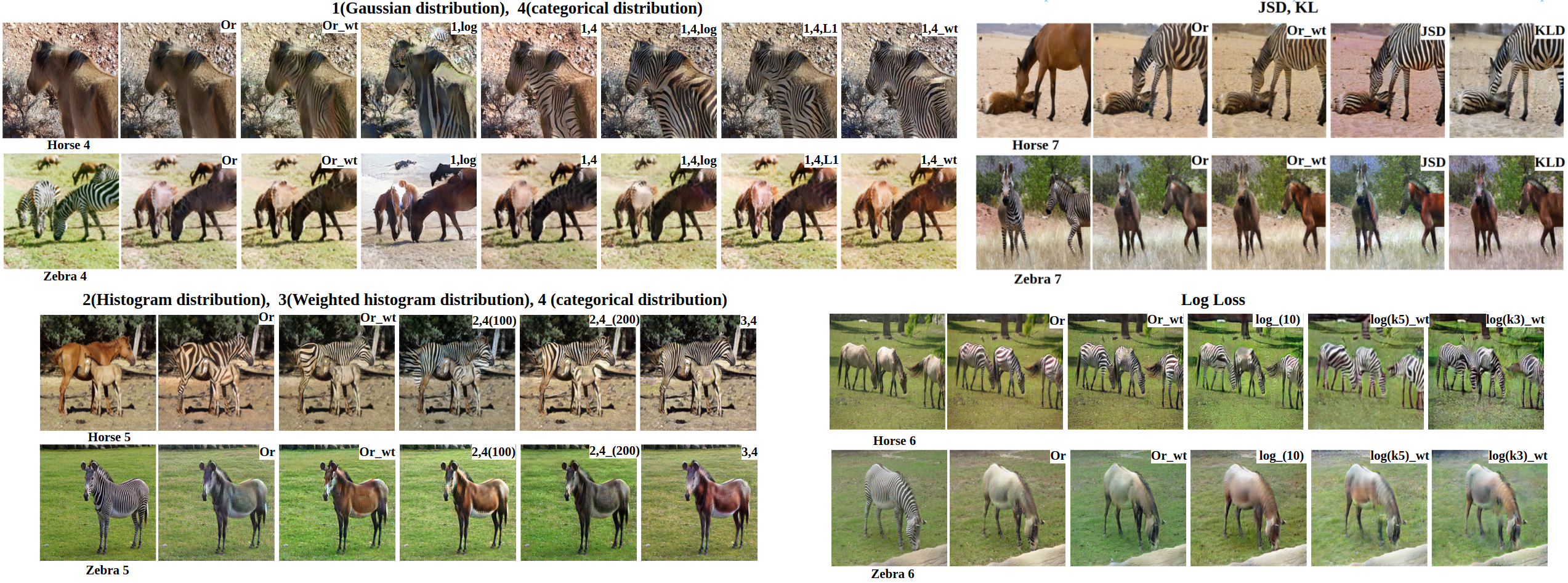}
    \caption{Comparison results of Latent representations learning categorized on the basis of similarity of different objectives of the experiments on Horse2Zebra data set for Horse to Zebra and vice versa translation with various combinations of frequency distributions and loss metrics. Frequency distributions: 1- Gaussian distribution (\textit{Gauss}), 2-Histogram (\textit{Hist}) 3-Weighted histogram (\textit{wtHist}) 4-Categorical distribution (\textit{Catg}) 5-Categorical patch distribution (\textit{CatgP}). Loss metrics: Divergence (jsd, kld), log}
    \label{fig:g_pdf}
\end{figure*}

\subsection{Effect of Loss Functions}
Traditionally, cycle-consistency loss employs L1 norm to  compute the pixel-wise absolute difference between reconstructed and original images. This encourages structural alignment by enforcing a direct correspondence in pixel intensities. However, the L1 loss promotes sparsity in the latent space by activating only a few dimensions for a given input, which may hinder the model’s ability to capture nuanced contextual and local neighborhood information. In contrast, the statistical divergence functions (KLD/JSD), have proven more effective in learning the underlying distribution of both local and global structures within the data. These divergence measures facilitate better alignment between the semantic features of the source and target domains, leading to enhanced image quality and sharper details, as reflected in the results shown in Figure~\ref{fig:h2z_wt_cyclegan_all}. Additionally, the use of logarithmic loss functions has demonstrated comparable performance, contributing to the preservation of fine details in the generated images. The integration of divergence and logarithmic losses, alongside frequency-based latent representation learning has improved the convergence behavior of our FD CycleGAN model. These enhancements have also mitigated mode collapse, resulting in a more diverse and stable generation of image samples across domains.

\subsection{Performance on Various Datasets}
Dataset diversity plays a pivotal role in evaluating the generalization and adaptability of image translation models. To this end, we have selected three visually and semantically distinct datasets—Horse2Zebra, Monet2Photo, and a Strike-off handwritten dataset—to rigorously assess our proposed framework. In the Horse2Zebra dataset, the model must learn intricate mappings of textures, colors, and structural contours between horse and zebra images. The FD CycleGAN, aided by learned latent representations and frequency-aware functions, effectively captures these domain-specific attributes and translates them into realistic outputs. The Monet2Photo dataset presents a more abstract challenge, requiring the preservation of artistic texture and color palettes across painting and real-world domains. Our model successfully retains the stylistic essence of Monet paintings while improving the fidelity of photorealistic reconstructions. This demonstrates its capacity to manage both stylistic abstraction and realistic rendering within a single framework. The Strike-off dataset introduces a unique scenario involving handwritten text with strike-through markings. While the original CycleGAN struggled to remove such artifacts due to limited representation learning, our proposed approach demonstrates strong capabilities in identifying and removing these strike-off patterns. The frequency-based latent features enable the model to learn subtle variations in pen strokes and ink intensity, resulting in clean, smooth, and strike-off-free handwriting generation. These results collectively underscore the versatility of the proposed ProbDist CycleGAN in handling a wide spectrum of image translation tasks—from natural scenes and artistic domains to document restoration—through robust frequency-aware representation learning and improved loss formulations.

\vspace{0.1cm}
\subsection{Overall Observations}

\vspace{0.1cm}

\textbf{Generalization and Adaptability.} One of the key takeaways of this study is the demonstrated generalizability of the Fd-CycleGAN across diverse datasets with varying domain characteristics. While traditional CycleGANs are often limited in their ability to adapt to non-homogeneous or complex image distributions, Fd-CycleGAN exhibits consistent translation performance across artistic (Monet2Photo), natural (Horse2Zebra), and structural (Strike-off handwritten) domains. This adaptability arises from the model’s capability to learn local and global data characteristics via frequency-aware latent space embedding.

\textbf{Latent Representations as a Semantic Bridge.} The proposed frequency distribution functions serve as a semantic bridge that not only improves visual fidelity but also preserves domain-specific content. This semantic coherence is critical for practical applications, particularly in domains like document image cleaning or artistic style translation, where minor deviations from structure or content can have significant implications.

\textbf{Avoidance of Mode Collapse.} The use of divergence-based loss functions and carefully constructed frequency distributions contributes to the model’s ability to avoid mode collapse. This stability is evident in the consistent diversity of generated outputs without sacrificing content or structural fidelity. The hybrid usage of Gaussian and categorical priors encourages exploration in the latent space, enabling the model to capture a wider spectrum of possible outputs.

\textbf{Limitations and Future Scope.} While Fd-CycleGAN demonstrates notable improvements, certain edge cases—such as extreme lighting variations or highly occluded input images—still pose challenges. Future work can explore integrating attention-based mechanisms with distribution learning to further enhance fine-grained feature localization. Moreover, frequency domain learning can be extended to video translation tasks or multimodal generation tasks by incorporating temporal consistency constraints.

\subsection{Comparison with Existing Approaches}

Table~\ref{tab:metrics1} presents a comprehensive comparison of our proposed \textit{Fd-CycleGAN} framework against several state-of-the-art I2I translation models for the \textit{Horse to Zebra} task at an object-level resolution of $256 \times 256$. Among the baseline methods, CycleGAN achieves a moderate balance across all metrics with a PSNR of 18.53 and SSIM of 0.67, while CUT~\cite{CUT2020contrastive} shows a lower PSNR and SSIM but a significantly improved FID score of 45.50, indicating better perceptual realism. CycleDiff, although a diffusion-based model, performs poorly in terms of both fidelity (PSNR of 11.51, SSIM of 0.21) and semantic alignment (CLIP score of 25.04), with a notably high FID of 347.27. CycleNet shows a PSNR of 20.42 but still lags in perceptual quality and semantic alignment. In contrast, our proposed \textit{Fd-CycleGAN} variants consistently outperform baseline models in most quality metrics. The \textbf{Fd-CycleGAN(L1)} variant achieves the highest PSNR (25.98) and SSIM (0.89), demonstrating superior reconstruction quality. Although the FID of this variant (81.67) is on par with CycleGAN, the CLIP score (29.13) shows better semantic preservation. Furthermore, the \textbf{Fd-CycleGAN(JSD)} and \textbf{Fd-CycleGAN(KLD)} variants yield significantly improved FID scores of 67.53 and 76.28 respectively, suggesting enhanced realism in the generated outputs. The \textbf{Fd-CycleGAN(Gauss+Catg)} and \textbf{Fd-CycleGAN(Hist+Catg)} models offer a favorable trade-off between perceptual and structural quality, maintaining PSNR values above 23 and SSIMs above 0.83. These results highlight the efficacy of frequency distribution-based constraints, especially when combined with local spatial features (category histograms or Gaussian priors).

Overall, the results confirm that incorporating frequency distribution-based losses in CycleGAN leads to notable improvements in fidelity, structure, and semantic consistency, validating the robustness of our proposed \textit{Fd-CycleGAN} framework across multiple evaluation criteria.

\section{Conclusion}\label{sec:conl}
The current study aimed to determine the effectiveness of latent representations in the task of I2I translation. We proposed \textit{LNE} and \textit{Fd} functions to learn the latent distributions of different domains to enhance CycleGAN learning. The combination of local neighborhood preservation with the Gaussian distribution function and global structural information with the categorical distribution function could capture the fine details of different domains while preserving the texture of the respective classes. The Histogram distribution effectively represented discrete data information of image pixels, which aided the translation of pixel-level information of the target distribution to the generated samples. Log loss produced similar results. The combination of these loss functions, along with the latent representations learning, was able to avoid mode collapse and increase the convergence of the model. Overall, this study strengthened the idea that the latent representation perspective improves the model's ability to learn semantic relationships between distributions to produce visually pleasing images.

\bibliographystyle{IEEEtran}  
\bibliography{refer}

\end{document}